\documentclass{article}
\usepackage{iclr2027/iclr2027_conference,times}

\usepackage{amsmath,amsfonts,bm}

\def\eqref#1{equation~\ref{#1}}

\def\1{\bm{1}}

\DeclareMathAlphabet{\mathsfit}{\encodingdefault}{\sfdefault}{m}{sl}
\SetMathAlphabet{\mathsfit}{bold}{\encodingdefault}{\sfdefault}{bx}{n}

\usepackage{hyperref}
\hypersetup{hidelinks}
\usepackage{url}
\usepackage{graphicx}
\usepackage{booktabs}
\usepackage{amsmath,amssymb,amsthm}
\usepackage{multirow}
\usepackage{longtable}
\usepackage{xcolor}
\usepackage{float}

\newcommand{\tis}{\textsc{Tis}}
\newcommand{\Tb}{T_{b}}

\title{Cool the Sampler, Not the Learner: Sampling\\ Temperature Moves the Staleness Cliff\\ of Importance-Corrected GRPO}

\author{Taiheng Pan\\ The University of Melbourne\\ \texttt{taiheng.pan@student.unimelb.edu.au}}

\iclrfinalcopy
\begin{document}
\maketitle

\begin{abstract}
Production RL for language models lets the sampler fall behind the learner and repairs the resulting mismatch with a truncated importance weight. We ask how long the sampler can go without a refresh under that correction, and find a cliff: on Qwen2.5-Math-1.5B and GSM8K, importance-corrected GRPO refreshed every 192 updates learns well for 180 steps and then degrades severely in all three data seeds before the refresh arrives. Published remedies for staleness act on the update; we act on the sampler instead. Decoupled cooling draws samples at temperature 0.8 while the learner, the reference model and the importance weights stay at temperature 1, with the behaviour probability recorded from the tempered distribution, so the learner's objective is unchanged. All corresponding cooled runs are stable, and the longer interval keeps what the short one delivered: at the same update budget, a cooled sampler refreshed every 192 steps matches an uncooled sampler refreshed every 96 at the end of training (0.857 for both) and averaged over it (0.79), whereas lowering the learning rate to a safe value ends 3--7 points lower. On Qwen2.5-Math-7B the degradation points at interval 192 predict that an interval of 144 is fatal without cooling and survivable with it; on two data seeds the uncooled runs degrade before their first refresh and the cooled runs pass it and end at 92--93\% against 68--81\%, with one cooled run degrading transiently late in the second cycle. The benefit has a window: at three times the safe interval and in a high-mismatch MATH setting cooling delays degradation without preventing it, stronger cooling is not better, and cooling without the correction collapses. Sampling temperature is a control on staleness tolerance, and temperature and refresh interval should be chosen together.
\end{abstract}

\section{Introduction}
\label{sec:intro}

\begin{figure}[t]
\centering
\includegraphics[width=\linewidth]{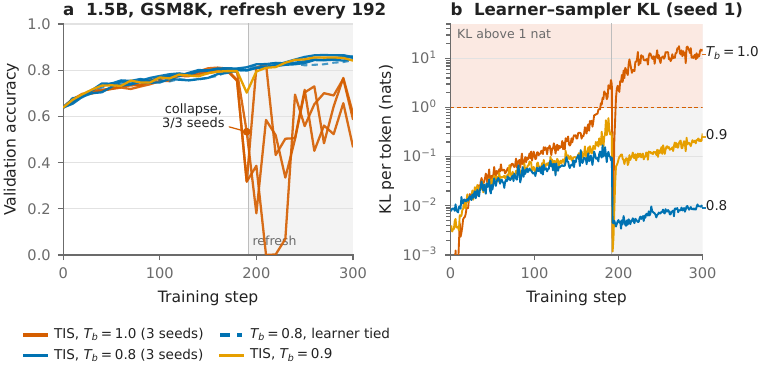}
\caption{\textbf{Importance-corrected GRPO has a finite safe staleness; cooling the sampler extends it.} Qwen2.5-Math-1.5B on GSM8K, sampler refreshed every 192 steps (grey line), learner at temperature 1, token-level \tis{} with cap 2 in all runs. (a) Validation accuracy (greedy, 1319 problems, every 10 steps). With $\Tb{=}1.0$ all three data seeds collapse at step 185--190 and oscillate for the rest of the run; with $\Tb{=}0.8$ all runs are stable (three data seeds with the learner at temperature 1, plus one run with the learner tied to 0.8); $\Tb{=}0.9$ dips once at step 190 and recovers. (b) Per-step KL between learner and sampler policies for the seed-1 runs of each temperature (log scale, dashed line at 1 nat per token). The KL accumulates within a refresh cycle, resets at the refresh, and grows faster the hotter the sampler; only the $\Tb{=}1$ run crosses one nat before the refresh.}
\label{fig:lag192}
\end{figure}

Production RL systems for language models let the sampler fall behind the learner, because generation is the bottleneck and waiting for fresh weights wastes it \citep{fu2025areal,rollflash2025,stalenessconstrained2026}. The price is a sampler whose distribution drifts from the learner's, and the standard remedy is a truncated importance weight on each token, \tis{} \citep{yao2025mismatch,sheng2024verl}. With the weight in place, how long can the sampler go without a refresh? Here is what that question looks like on Qwen2.5-Math-1.5B: three data seeds, importance-corrected GRPO, sampler refreshed every 192 steps. All three climb to about 0.81 on GSM8K. All three fall within the same ten steps, at step 185--190, before the refresh arrives, and end at 0.61, 0.59 and 0.47 (Figure~\ref{fig:lag192}). Validation was healthy through the first 150 steps. We call this the \emph{staleness cliff}: a refresh interval can look safe for most of a cycle and fail before the cycle ends.

The obvious way to buy more tolerance is to make the learner more conservative: a smaller step, a bound on the weights, a veto on suspicious samples, a re-targeted proximal term \citep{zheng2026m2po,huang2026vcpo,mugrpo2026,a3po2025,gac2026}, with a scaling law for how the collapse time shrinks with staleness times learning rate \citep{song2026staleness}. All of these act on the update. But the importance weight only decides how samples are used once they exist; the sampler decides which samples exist, and with them how fast the learner's probabilities on those samples drift from the sampler's. That leaves a second question, the one this paper asks: with the correction rule fixed, can the data be generated so that a longer refresh interval stays usable?

The sampler has one knob for that, its temperature, and we turn nothing else. The sampler draws at $\Tb{=}0.8$; the learner, the reference model, the KL penalty and the importance weights are all evaluated at temperature 1, and the behaviour probability recorded for the weight is that of the tempered distribution that produced the token. The learner's objective is untouched; it is fed a sharper proposal whose probabilities are recorded correctly. We call this \emph{decoupled cooling}. It is a two-line change to verl.

Under it the cliff moves. At a refresh interval of 192 the cooled 1.5B runs are stable in four of four, with the KL never above 0.2 nats; at $\Tb{=}0.9$, halfway between, the run dips once at step 190 and recovers, so the cliff is a threshold in the growth of the mismatch, and temperature sets the rate. What matters more is what the longer interval keeps. On the same seeds and the same 300-update budget, an uncooled sampler refreshed every 96 steps and a cooled sampler refreshed every 192 end at the same accuracy, 0.857, and average the same accuracy over the run, 0.790 against 0.787 (Table~\ref{tab:budget}). Lowering the learning rate to a safe value also avoids the cliff but ends 3--7 points lower in the same budget; the closest update-side stabiliser, $\mu$-GRPO, ends below cooling on both seeds and gains nothing from cooling on top. Cooling buys the longer interval without giving up the learning the short one delivered.

The finding can guide the next choice, and the 7B model let us test that as a prediction. Qwen2.5-Math-7B reaches its cliff sooner: at a refresh interval of 192 the uncooled run degrades at about 120 steps of staleness and the cooled run at about 180 (Figure~\ref{fig:7b}b). A refresh interval of 144 should therefore be fatal without cooling and survivable with it. On two data seeds it was: the uncooled runs degrade severely before their first refresh, the cooled runs reach it intact, keep learning, and end at 0.92 and 0.93 against 0.68 and 0.81 (Figure~\ref{fig:7b}a). On one seed the cooled run degrades transiently late in the second cycle and recovers at the next refresh: the safe interval depends on the training state, and passing one refresh is not a permanent guarantee.

The cliff can be moved, not removed. At three times the safe interval every temperature degrades before or just after the refresh, in an order that is not monotone in temperature; on MATH with a doubled learning rate cooling delays and softens the collapse but does not prevent it; $0.6$ is worse than $0.8$ wherever we tried it; cooling without the correction only postpones collapse; heating the sampler when groups saturate collapses it. The effect has a temperature window and depends on task and training stage, which is the point: temperature and refresh interval are one design decision, not two.

Our contributions are: (i) the staleness cliff, a reproducible failure of importance-corrected GRPO at moderate refresh intervals on two model sizes, with its signature in the within-cycle KL; (ii) decoupled cooling, a sampler-side change that moves the cliff without changing the learner's objective, verified across seeds on the 1.5B model and by a prediction on the 7B model, and which keeps the learning progress of the short refresh interval with a third of the refreshes, and learns more in the same budget than a safe lower learning rate; (iii) the limits: where the cliff cannot be moved, which related interventions do not help, and why the usual diagnostics are not alarms, all reported under fixed degradation criteria. The change is a two-line patch to verl on a part of the pipeline the objective-side stabilisers above do not touch; against one of them, $\mu$-GRPO, it is as stable and more accurate in the same budget on both tested seeds, and the two do not add (Section~\ref{sec:update}).

\section{Setup}
\label{sec:setup}

\paragraph{Training.} We use GRPO \citep{shao2024deepseekmath} as implemented in verl \citep{sheng2024verl} with vLLM~0.11 as the sampler \citep{kwon2023vllm}. Each step draws 16 prompts and 8 samples per prompt (512 prompt tokens, 512 response tokens on GSM8K, 1024 on MATH), computes group-normalised advantages from a rule-based correctness reward, and takes one optimiser step on the full batch (one PPO epoch, so a ratio to the epoch-start policy is identically one and its clipping inert; $\mu$-GRPO in Section~\ref{sec:update} takes its ratio to the frozen sampler instead). Learning rate $2{\times}10^{-6}$ unless stated, a low-variance KL penalty to the initial policy with coefficient $10^{-3}$, no entropy bonus, four GPUs per run. Models are Qwen2.5-Math-1.5B and Qwen2.5-Math-7B \citep{yang2024qwen25math}. Tasks are GSM8K \citep{cobbe2021gsm8k} (train split; validation on the 1319-problem test split) and MATH levels 3--5 \citep{hendrycks2021math} (5586 training problems; validation on MATH-500 \citep{lightman2024verify}). Validation is greedy and is run every 10 steps on GSM8K and every 20 on MATH, always with fresh weights.

\paragraph{Staleness.} We emulate the decoupled-sampler regime by refreshing the sampler's weights only every $N$ learner steps; between refreshes the sampler is frozen, so the first $N$ steps of every run sample from the initial policy and the staleness at step $t$ is $t \bmod N$. Refresh intervals are $N \in \{96, 144, 192, 288\}$ over 300 steps, so a run covers two or three refreshes. This is the ``few large stages'' regime studied by \citet{mugrpo2026}, with the sampler's lag made explicit.

\paragraph{Correction.} All runs labelled \tis{} use verl's token-level truncated importance weight $\min(\pi_\theta(a_t\mid s_t)/q(a_t\mid s_t),\,2)$ between the learner $\pi_\theta$ and the recorded behaviour probability $q$ of the sampler, applied as a fixed per-token coefficient on the policy-gradient loss \citep{yao2025mismatch}. The weight corrects the gradient on the responses that were drawn, not which responses are drawn: every group's normalised advantage is set by the sampling distribution, and the truncation caps the mismatch the weight can undo. Runs labelled ``no IS'' use plain GRPO with no correction.

\paragraph{Sampling temperature, decoupled.} The intervention is a single knob. The sampler draws from $q_{\Tb}(a\mid s)\propto \pi_{\text{sampler}}(a\mid s)^{1/\Tb}$ with $\Tb\le 1$, and the behaviour probability recorded for the importance weight is $q_{\Tb}$ itself (vLLM's \texttt{processed\_logprobs}; the default \texttt{raw\_logprobs} would record the untempered probability and silently misstate the weight). The learner, the reference model and the KL penalty are evaluated at temperature 1, so the surrogate objective and the learner's log-probabilities are unchanged: cooling changes which samples are proposed and how they are weighted, not what the learner is asked to maximise. With weights truncated at 2, which tokens hit the cap depends on the proposal, so a cooled sampler changes the bias of the estimator as well as its variance; what stays fixed is the target. We call this the decoupled setting. In the tied setting the learner is also evaluated at $\Tb$, which changes the objective; we include it as a control.

\paragraph{Degradation criteria.} All curves are reported in full; to summarise them we use fixed criteria. The \emph{onset} of degradation is the first validation point at or below the running maximum minus $0.10$ (after step 30); \emph{severe} degradation is at or below the maximum minus $0.15$; a degradation is \emph{transient} if the curve returns to within $0.05$ of the previous maximum within 20 steps and \emph{sustained} otherwise. Table~\ref{tab:onsets} lists these, and the first step at which the learner--sampler KL exceeds one nat, for every degrading run.

\section{The staleness cliff, and moving it}
\label{sec:main}

\subsection{1.5B: no cliff at 96, a cliff at 192, no cliff at 192 when cooled}

At this scale the cliff sits between refresh intervals 96 and 192, and cooling moves it past 192. Refreshing every 96 steps is safe for \tis{} at temperature 1: three data seeds end at 0.861, 0.860 and 0.864. Cooling costs nothing here: with $\Tb{=}0.8$ the three seeds end at 0.856, 0.860 and 0.865, and the runs reach 0.85 about 45 steps earlier on average.

Refreshing every 192 steps is not safe (Figure~\ref{fig:lag192}a). The three $\Tb{=}1$ seeds reach 0.80--0.81 by step 160--170 and all reach severe degradation at step 190 (0.53, 0.32 and 0.32), five to twenty steps after their KL crosses one nat (at steps 179, 171 and 181). The refresh at 192 loads the degraded learner into the sampler and the runs oscillate to the end (0.61, 0.59 and 0.47; KL maxima 9--18 nats). With $\Tb{=}0.8$ nothing of the sort happens: the three decoupled seeds end at 0.848, 0.857 and 0.859, the tied run at 0.843, and the KL never exceeds 0.20 nats. At $\Tb{=}0.9$ the run dips once to 0.70 at step 190 and recovers to 0.855 (KL maximum 0.61).

Figure~\ref{fig:lag192}b shows why. Within a cycle the KL between learner and sampler grows roughly exponentially, resets at the refresh, and grows faster the hotter the sampler. At $\Tb{=}1$ it reaches one nat at step 179; at $\Tb{=}0.9$ it reaches 0.6 and turns over; at $\Tb{=}0.8$ it stays below 0.2. Every collapse of a fixed-temperature corrected run here is preceded or accompanied by the KL crossing one nat, at no fixed lead; the heated controller run of Section~\ref{sec:boundary} is the one collapse without a crossing. Cooling does not change the correction; it keeps the mismatch smaller for longer. The practical reading is that cooling lets the refresh interval double at the end quality of the short one: on seeds 43 and 44 the last-five mean is 0.857 at $N{=}96$ uncooled and 0.857 at $N{=}192$ cooled, and the whole-run mean 0.790 against 0.787 (Table~\ref{tab:budget}).

\subsection{The mismatch, across every run}
\label{sec:mech}

\begin{figure}[t]
\centering
\includegraphics[width=\linewidth]{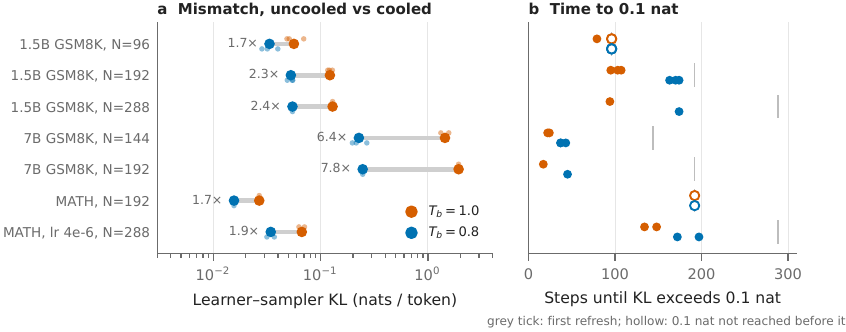}
\caption{\textbf{Cooling lowers the learner--sampler mismatch in every setting.} (a) KL at a fixed step of the first cycle (110, or 60 where the refresh is at 96), uncooled ($\Tb{=}1$, orange) against cooled ($0.8$, blue), one row per model, task and refresh interval; large dots are seed means, small dots single runs, the label is their ratio. (b) First step at which the KL of the same runs exceeds $0.1$ nat per token; the grey tick marks the first refresh, and hollow markers did not reach $0.1$ nat before it.}
\label{fig:mech}
\end{figure}

Cooling does not slow the learner down; it changes what the learner has to correct. The picture in Figure~\ref{fig:lag192}b is not specific to those three runs (Figure~\ref{fig:mech}). We read the learner--sampler KL of every importance-corrected run at a fixed step of the first cycle (110, or 60 where the refresh is at 96), before any run has degraded, and record the first step at which it exceeds $0.1$ nat. Cooling lowers the mismatch in every setting, by 1.7 to 7.8 times at the same step: at step 110 the cooled 1.5B runs sit at 0.05--0.06 nats against 0.12--0.13 for $\Tb{=}1$ on GSM8K, and the cooled 7B runs at 0.20--0.27 against 1.3--1.9. The time to reach $0.1$ nat lengthens accordingly (1.5B: about 100 to 150--175 steps; 7B: about 20 to 40). The same reading says why cooling below $0.8$ buys nothing (Section~\ref{sec:boundary}). Extrapolating the KL growth of steps 20--60 to the refresh does not separate runs that later degrade from runs that do not, so the mismatch describes what cooling changes and does not forecast the cliff.

Cooling is also not a disguised reduction of the learning rate. The learner's KL to its initial policy at step 90 is 2--8 times \emph{smaller} in the cooled runs (0.004--0.007 against 0.010--0.015 nats at 1.5B; 0.005--0.011 against 0.029--0.043 at 7B) at equal or higher validation accuracy: the same progress with less movement. Section~\ref{sec:lr} tests the learning-rate reading directly.

\subsection{7B: the cliff arrives earlier, and a prediction about moving it}

\begin{figure}[t]
\centering
\includegraphics[width=\linewidth]{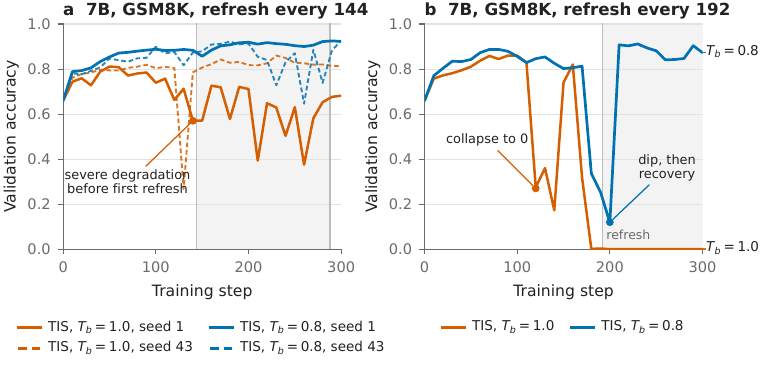}
\caption{\textbf{Qwen2.5-Math-7B on GSM8K.} (a) Refresh every 144 steps, chosen before the runs from the two runs in (b); solid lines data seed 1, dashed seed 43. The uncooled runs degrade severely before the first refresh on both seeds; the cooled runs have no degradation before it, keep improving after it, and end at 0.92--0.93 (all realizations in Appendix~\ref{app:7b}). (b) Refresh every 192 steps: the uncooled run collapses to zero; the cooled run degrades 60 steps later and recovers after the refresh.}
\label{fig:7b}
\end{figure}

The 7B model sharpens faster and hits the limit sooner (Figure~\ref{fig:7b}). At $N{=}192$ the uncooled run reaches 0.86 by step 90, crosses one nat at step 92, drops to 0.27 at step 120 and to zero at step 180, and stays there. The cooled run reaches 0.89 by step 70, degrades severely at step 180 (0.34), recovers after the refresh to 0.91 and ends at 0.87. Cooling moved the onset from step 120 to 180 and turned a permanent collapse into a recoverable one, but at $N{=}192$ it did not prevent the first-cycle degradation.

These two runs turn the claim into a prediction. The uncooled 7B model degrades at about 120 steps of staleness and the cooled one at about 180, so a refresh interval of 144 should sit between the two: the uncooled run should fall before its first refresh, and the cooled run should reach it intact. We ran both arms for 300 steps, through refreshes at 144 and 288, on two data seeds, and the prediction held on both. Before its first refresh the uncooled run fell each time, on seed 1 at step 140 for good and on seed 43 at step 130 with a recovery ten steps later; what the refresh then loaded into the sampler was a degenerate policy on both seeds, which for the whole second cycle ran 84--95\% of its responses to the length limit and kept the training reward below 0.2, whether or not greedy validation showed it. The cooled run reached the refresh intact on both seeds and in a third, interrupted, realization, and kept improving after it, to 0.91--0.92 by step 200, with a healthy sampler. The second cycle is where the seeds part: seed 1 passed the second refresh without an event; seed 43 oscillated from step 230, with the KL above one nat and the entropy rising, and recovered at the refresh at 288. Cooling moved the first cliff past 144 on both seeds and gave the higher final accuracy (0.92 and 0.93 against 0.68 and 0.81) and the higher whole-run mean (0.880 and 0.841 against 0.673 and 0.792) on both; the last-five mean is higher on seed 1 only (0.915 against 0.593; 0.816 against 0.818 on seed 43). It did not make the second cycle uniformly safe (Appendix~\ref{app:7b}).

\section{Three objections, and where the cliff cannot be moved}
\label{sec:boundary}

Three objections come first, each with a direct test; then the settings where the cliff cannot be moved.

\begin{figure}[t]
\centering
\includegraphics[width=\linewidth]{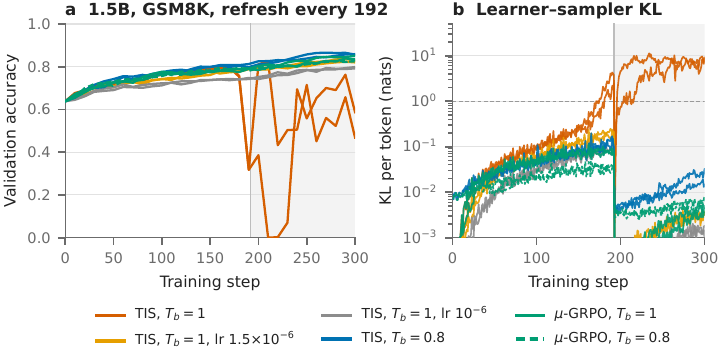}
\caption{\textbf{Lowering the learning rate or an update-side stabiliser ($\mu$-GRPO) also avoids the cliff; cooling learns more, and does not stack with $\mu$-GRPO.} 1.5B, GSM8K, refresh every 192, data seeds 43 and 44 for every arm. (a) Validation accuracy: \tis{} at $\Tb{=}1$ with learning rate $2{\times}10^{-6}$ collapses; at $1.5{\times}10^{-6}$ and $10^{-6}$ it is stable but ends at 0.82--0.83 and 0.79--0.80; the cooled sampler at $2{\times}10^{-6}$ is stable and ends at 0.86. $\mu$-GRPO (green; solid $\Tb{=}1$, dashed $0.8$) is stable in both seeds (last-five mean 0.83 and 0.85 alone, 0.82 with the cooled sampler). (b) Learner--sampler KL of the same runs.}
\label{fig:lr}
\end{figure}

\paragraph{Is cooling a smaller learning rate in disguise?}
\label{sec:lr}
Could a lower learning rate give the same stability, since cooling slows the learner's drift (Section~\ref{sec:mech})? \tis{} at $\Tb{=}1$ with the learning rate lowered to $1.5{\times}10^{-6}$ and $10^{-6}$, same seeds and refresh interval, does avoid the cliff: all four runs are stable (KL maxima 0.24--0.25 and 0.11--0.12 nats; Figure~\ref{fig:lr}). It also learns less in the same budget: over the last five validations cooling at $2{\times}10^{-6}$ ends 3.4 points above the best tested lower learning rate and 6.8 above the lower one (Table~\ref{tab:budget}), and the cooled runs lead throughout the second cycle in both seeds. Within the tested learning rates and a 300-update budget, cooling gives the better stability--progress trade-off.

\paragraph{Is an update-side stabiliser enough?}
\label{sec:update}
Does a stabiliser acting on the update already do what cooling does? We compare against the $\mu$-GRPO update rule \citep{mugrpo2026} under our common recipe: the ratio is taken to the frozen sampler's log-probabilities, clipped to $[0,5]$, and a negative-advantage response containing a token with ratio below $10^{-4}$ is vetoed; no additional truncated importance multiplier is applied (port details in Appendix~\ref{app:impl}). At $N{=}192$ the rule is active (3--5\% of responses vetoed late in the first cycle) and both seeds are stable. Over the last five validations cooled \tis{} is ahead of $\mu$-GRPO by 3.5 and 0.6 points on the two seeds (Table~\ref{tab:budget}, Figure~\ref{fig:lr}). Cooling the sampler under $\mu$-GRPO does not add: the mismatch and the veto rate fall, but accuracy does not rise. Under this recipe and budget, cooling is a competitive alternative to this update-side stabiliser and does not add to it.

\begin{table}[t]
\centering\small
\caption{\textbf{Same budget, same seeds.} 1.5B, GSM8K, 300 updates, learning rate $2{\times}10^{-6}$ unless stated, data seeds 43 and 44: severe-degradation count, the end metric (last-five mean, steps 260--300) per seed and averaged, and the whole-run metric (trapezoid mean of the 31 validations, averaged over seeds). Cooling doubles the refresh interval at the end and whole-run quality of the short one; the alternatives avoid the cliff at a cost on both. Updates and prompt budget are identical across rows; fewer refreshes pay off where the refresh is the bottleneck.}
\label{tab:budget}
\vspace{2pt}
\begin{tabular}{llccccccc}
\toprule
Objective & $\Tb$ & $N$ & Refreshes & Degraded & \multicolumn{3}{c}{Last five: 43 / 44 / mean} & Whole run \\
\midrule
\tis{} & 1.0 & 96 & 3 & 0/2 & 0.854 & 0.860 & 0.857 & 0.790 \\
\tis{} & 0.8 & 96 & 3 & 0/2 & 0.860 & 0.862 & 0.861 & 0.800 \\
\tis{} & 1.0 & 192 & 1 & 2/2 & 0.679 & 0.532 & 0.605 & 0.654 \\
\tis{} & 0.8 & 192 & 1 & 0/2 & 0.863 & 0.852 & 0.857 & 0.787 \\
\tis{}, lr $1.5{\times}10^{-6}$ & 1.0 & 192 & 1 & 0/2 & 0.818 & 0.830 & 0.824 & 0.761 \\
\tis{}, lr $10^{-6}$ & 1.0 & 192 & 1 & 0/2 & 0.790 & 0.789 & 0.789 & 0.734 \\
$\mu$-GRPO & 1.0 & 192 & 1 & 0/2 & 0.829 & 0.846 & 0.837 & 0.772 \\
$\mu$-GRPO & 0.8 & 192 & 1 & 0/2 & 0.824 & 0.824 & 0.824 & 0.770 \\
\bottomrule
\end{tabular}
\end{table}

\begin{figure}[t]
\centering
\includegraphics[width=\linewidth]{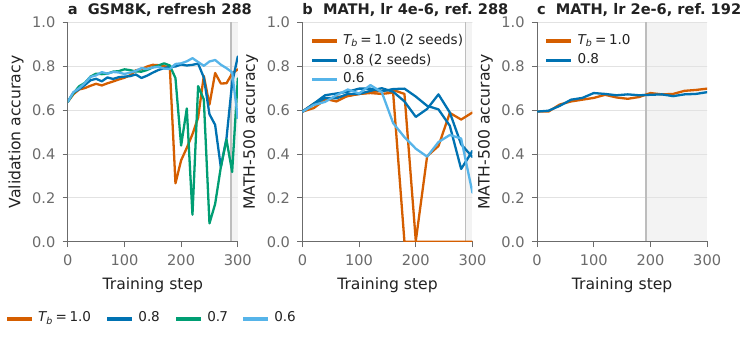}
\caption{\textbf{Boundaries.} (a) 1.5B, GSM8K, refresh every 288: every temperature degrades before or just after the refresh and the onset is not monotone in temperature. (b) 1.5B, MATH levels 3--5 with learning rate $4{\times}10^{-6}$ and refresh every 288 (validation on MATH-500): uncooled runs collapse in both seeds; cooled runs degrade gradually and later; $\Tb{=}0.6$ is worse than $0.8$. (c) MATH at the standard learning rate and refresh 192: no run degrades and cooling is neutral.}
\label{fig:boundary}
\end{figure}

\paragraph{Is colder better?} No. At $N{=}288$ the onset is not monotone in temperature ($0.8$ degrades at step 250, $0.7$ at 190--200, $0.6$ at 300), and in the high-mismatch MATH setting $0.6$ degrades earlier than the uncooled run (Figure~\ref{fig:boundary}). Section~\ref{sec:mech} says why: below $0.8$ the mismatch at a fixed step stops falling (0.055, 0.060 and 0.083 at step 110 for $0.8$, $0.7$ and $0.6$), so a colder sampler pays the diversity cost and buys no further protection. The window is near $0.8$ at these settings; Section~\ref{sec:discussion} reconciles this with the on-policy finding that $0.6$ is unstable and $1.2$ safe \citep{liu2025prorl}.

\paragraph{Larger staleness.} At $N{=}288$ on the 1.5B model, no temperature we tried is stable through 300 steps (Figure~\ref{fig:boundary}a). The uncooled run degrades at step 190, as at $N{=}192$: the collapse is a property of the staleness reached, not of the refresh boundary. Cooling to $0.8$ moves the onset to step 250 and colder settings do not move it further. Single runs vary by tens of steps; $0.8$ roughly doubles the safe interval at these settings, and no temperature triples it.

\paragraph{A harder task.} On MATH the 1.5B model gains slowly (0.59 to 0.68--0.70 in 300 steps), sharpens little, and the KL stays below 0.1 nats at $N{=}192$ for both temperatures; neither run degrades, and cooling neither helps nor costs (0.682 versus 0.698, within noise). To reach the high-mismatch regime on MATH we double the learning rate and refresh every 288 steps, since instability scales with staleness times learning rate \citep{song2026staleness}. The uncooled runs then collapse in both seeds (onsets 180 and 200); the cooled runs degrade gradually, with the KL crossing delayed by 76 and 43 steps, and end at 0.41 and 0.39 (Figure~\ref{fig:boundary}b). Cooling changes the failure mode and delays it here, but does not prevent it.

\begin{figure}[t]
\centering
\includegraphics[width=\linewidth]{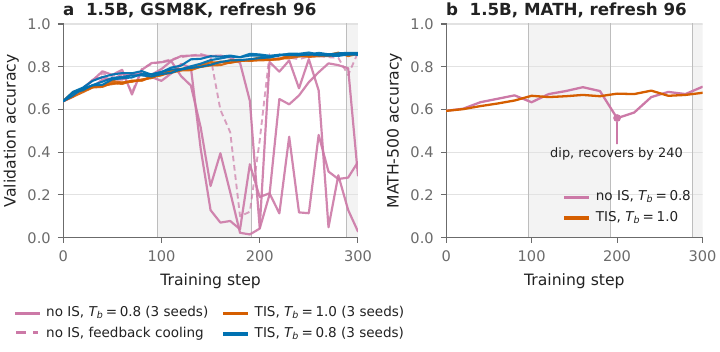}
\caption{\textbf{Cooling without the importance correction only delays collapse.} (a) 1.5B, GSM8K, refresh every 96. Plain GRPO with a cooled sampler and no correction (three seeds, plus a variant that cools further whenever the KL grows) collapses in the second cycle; the same cooled sampler with \tis{} is stable, as is uncooled \tis{}. (b) 1.5B, MATH, refresh every 96: the uncorrected cooled run dips at step 200 and recovers by step 240.}
\label{fig:noIS}
\end{figure}

\paragraph{Cooling without the correction.} The correction is necessary. Plain GRPO with no importance weight and a cooled sampler ($\Tb{=}0.8$, learner at 1, $N{=}96$) looks healthy for a hundred steps, reaching 0.81--0.85 by step 110--140, and then collapses in all three seeds, to 0.03, 0.35 and 0.29 (Figure~\ref{fig:noIS}a). The refresh at 96 does not save it; the same configuration with \tis{} is one of the stable arms of Section~\ref{sec:main}. Cooling and correction are complementary: cooling slows the growth of the mismatch, the correction absorbs what remains, and either alone fails. On MATH the uncorrected cooled run dips from 0.70 to 0.56 at step 200 and recovers by 240 (Figure~\ref{fig:noIS}b).

\paragraph{Adaptive temperature.} Reheating once groups saturate does not help: a saturation-triggered controller and a fixed 0.8-to-1.0 schedule end within 0.01 of fixed cooling at $N{=}96$, and at $N{=}192$ the controller heated to 1.05 in the second cycle and collapsed to 0.19 (Appendix~\ref{app:adaptive}): heating is the wrong direction under staleness.

\paragraph{Diagnostics are not early warnings.} Every collapse is accompanied by a KL excursion, an entropy that collapses (to 0.03--0.05 nats, in the uncorrected and 7B uncooled runs) or explodes (to 3--7 nats, in the cooled high-mismatch runs), and gradient spikes. The KL crossing one nat is nevertheless not an alarm (Table~\ref{tab:onsets}): it leads the onset by 9--42 steps in the uncooled \tis{} runs, by 79 steps in one cooled run, follows it by 16--27 steps in the $\Tb{=}0.7$ runs, and never happens in the collapsing controller run. We report these quantities as descriptions of the degradation, and the validation curve as the measurement.

\begin{table}[t]
\caption{Degradation events under the fixed criteria of Section~\ref{sec:setup}. Onset: first validation point $\le$ running max $-0.10$; KL$>$1: first step with learner--sampler KL above one nat per token; lead $=$ onset $-$ KL$>$1 (positive: KL crossed first). Peak: highest validation point of the run. Runs with no onset within 300 steps are listed as stable. Complete per-run numbers are in Appendix~\ref{app:ledger}.}
\label{tab:onsets}
\centering
\resizebox{\linewidth}{!}{%
\begin{tabular}{llllrrrrl}
\toprule
Model & Task & $N$ & Arm & Peak & Onset & KL$>$1 & Lead & Outcome \\
\midrule
1.5B & GSM8K & 192 & \tis, $\Tb{=}1$, 3 seeds & 0.81/0.80/0.83 & 190/190/190 & 179/171/181 & +11/+19/+9 & sustained, end 0.61/0.59/0.47 \\
1.5B & GSM8K & 192 & \tis, $\Tb{=}0.8$, 4 runs & 0.85--0.87 & -- & -- & -- & stable, end 0.84--0.86 \\
1.5B & GSM8K & 192 & \tis, $\Tb{=}0.9$ & 0.86 & -- & -- & -- & one transient dip at 190, end 0.84 \\
1.5B & GSM8K & 192 & \tis{} + controller & 0.84 & 300 & -- & -- & sustained, end 0.19 (KL max 0.43) \\
1.5B & GSM8K & 288 & \tis, $\Tb{=}1$ & 0.81 & 190 & 179 & +11 & recovers after refresh, end 0.79 \\
1.5B & GSM8K & 288 & \tis, $\Tb{=}0.8$ & 0.85 & 250 & 245 & +5 & recovers after refresh, end 0.85 \\
1.5B & GSM8K & 288 & \tis, $\Tb{=}0.7$, 2 runs & 0.81/0.83 & 200/190 & 216/217 & $-16$/$-27$ & sustained \\
1.5B & GSM8K & 288 & \tis, $\Tb{=}0.6$ & 0.84 & 300 & 221 & +79 & slow decline, end 0.56 \\
1.5B & GSM8K & 96 & no IS, $\Tb{=}0.8$, 3 seeds & 0.85/0.81/0.83 & 190/140/130 & 184/93/133 & +6/+47/$-3$ & sustained, end 0.03/0.35/0.29 \\
1.5B & MATH & 288$^\dagger$ & \tis, $\Tb{=}1$, 2 seeds & 0.68/0.69 & 180/200 & 171/172 & +9/+28 & sustained, end 0.00/0.59 \\
1.5B & MATH & 288$^\dagger$ & \tis, $\Tb{=}0.8$, 2 seeds & 0.70/0.70 & 260/200 & 247/215 & +13/$-15$ & gradual, end 0.41/0.39 \\
1.5B & MATH & 288$^\dagger$ & \tis, $\Tb{=}0.6$ & 0.71 & 160 & 164 & $-4$ & sustained, end 0.22 \\
7B & GSM8K & 192 & \tis, $\Tb{=}1$ & 0.86 & 120 & 92 & +28 & sustained, end 0.00 \\
7B & GSM8K & 192 & \tis, $\Tb{=}0.8$ & 0.89 & 180 & 148 & +32 & recovers after refresh, end 0.87 \\
7B & GSM8K & 144 & \tis, $\Tb{=}1$, 2 seeds & 0.81/0.86 & 120/130 & 98/88 & +22/+42 & sustained, end 0.68 / transient, end 0.81 \\
7B & GSM8K & 144 & \tis, $\Tb{=}0.8$, 2 seeds & 0.93/0.93 & --/230 & --/232 & --/$-2$ & stable, end 0.92 / transient in cycle 2, end 0.93 \\
\bottomrule
\end{tabular}}
\\[2pt]{\footnotesize $^\dagger$ learning rate $4{\times}10^{-6}$. The onset criterion flags a spurious early dip (step 70) in one uncorrected run; the severe-degradation step (190) is reported instead.}
\end{table}

\section{Related work}
\label{sec:related}

\paragraph{Staleness and its correction.} The mismatch between an inference-engine sampler and the learner, even at zero lag, was identified by \citet{yao2025mismatch}, who introduced the truncated token-level importance weight used here. \citet{song2026staleness} derive a bias of order staleness times learning rate and a collapse-time law in accumulated drift; our uncooled results agree, and our cooled results add a third axis. \citet{zheng2026m2po} document the same late collapse under stale data and constrain the second moment of the weights; \citet{huang2026vcpo} scale the step by the effective sample size; \citet{mugrpo2026} train in a few large stages, as we do, and stabilise with relaxed clipping and a negative-advantage veto; \citet{a3po2025,gac2026} adjust the proximal target or the gradient direction. All of these act on the update; ours acts on the proposal and leaves the objective untouched (Section~\ref{sec:update} compares it with $\mu$-GRPO and tests the combination).

\paragraph{Temperature in on-policy RLVR.} Sampling temperature is usually discussed as an exploration control: \citet{liu2025prorl} report that on-policy training at $\Tb{=}0.6$ destabilises early while $1.2$ trains more slowly but more stably; \citet{multitemp2025} vary temperature per token and per rollout; \citet{tempmeta2026,lookinward2026} learn a temperature policy; \citet{tsopsd2026} distil a heated self-teacher to undo entropy collapse. All have sampler and learner at one policy, where cooling costs only diversity; Section~\ref{sec:discussion} reconciles them with the stale case.

\paragraph{Systems.} Asynchronous trainers bound staleness by construction \citep{fu2025areal,rollflash2025,stalenessconstrained2026}; our result gives them a cheap way to widen that bound.

\section{Discussion}
\label{sec:discussion}

\paragraph{What cooling does.} A sharper sampler proposes responses the learner already rates highly, on which the learner--sampler ratio drifts more slowly than on the low-probability tokens a hotter sampler adds. The effect scales with the mismatch: nothing to fix at $N{=}96$, the difference between collapse and stability at $N{=}192$, outgrown at $N{=}288$.

\paragraph{Reconciling with on-policy results.} On-policy, \citet{liu2025prorl} find $\Tb{=}0.6$ unstable and $1.2$ safe; under staleness we find the opposite ordering between $1.0$ and $0.8$, and heating collapsing. The two orderings describe two costs, diversity starvation when cold and uncorrectable drift when hot and stale, so changing the proposal helps or hurts depending on staleness. No universal temperature follows; what does is that temperature and refresh interval are one design decision, with the window near $0.8$ at these settings.

\paragraph{Limitations.} All runs use one model family, two mathematical tasks and 300 steps; the 7B evidence is two data seeds per arm. We emulate staleness with a fixed refresh interval; a production queue's lag is a distribution whose tail matters.

\section{Conclusion}
Importance-corrected GRPO tolerates staleness only up to a point, reached at moderate refresh intervals on a 1.5B and a 7B model. The correction cannot be asked to fix data it was never designed for; the sampler can be asked to produce data it can fix. The sampler's proposal distribution sets, in part, how long the learner keeps learning between refreshes, and cooling it extends that interval while keeping the learning progress of the short one at a fixed update budget, within a temperature window and with a dependence on task and training stage that are as clear as the gain. Sampling temperature and refresh interval are one design decision, and the fix is one line.

\subsection*{Reproducibility statement}
Section~\ref{sec:setup} gives every hyperparameter, the staleness emulation, the correction and the temperature decoupling; Appendix~\ref{app:impl} describes the three patches to verl that implement them, and Appendix~\ref{app:ledger} lists every run with its settings and outcome. The training logs of all runs, the validation curves, and the scripts that produce every figure and table from those logs will be released.

\bibliography{references,references_old_san}

\begin{thebibliography}{22}
\providecommand{\natexlab}[1]{#1}
\providecommand{\url}[1]{\texttt{#1}}
\expandafter\ifx\csname urlstyle\endcsname\relax
  \providecommand{\doi}[1]{doi: #1}\else
  \providecommand{\doi}{doi: \begingroup \urlstyle{rm}\Url}\fi

\bibitem[Cobbe et~al.(2021)Cobbe, Kosaraju, Bavarian, Chen, Jun, Kaiser, Plappert, Tworek, Hilton, Nakano, Hesse, and Schulman]{cobbe2021gsm8k}
Karl Cobbe, Vineet Kosaraju, Mohammad Bavarian, Mark Chen, Heewoo Jun, Lukasz Kaiser, Matthias Plappert, Jerry Tworek, Jacob Hilton, Reiichiro Nakano, Christopher Hesse, and John Schulman.
\newblock Training verifiers to solve math word problems.
\newblock \emph{arXiv preprint arXiv:2110.14168}, 2021.

\bibitem[Dang et~al.(2026)Dang, Lan, Wan, Zhao, and Lu]{tempmeta2026}
Haoran Dang, Cuiling Lan, Hai Wan, Xibin Zhao, and Yan Lu.
\newblock Temperature as a meta-policy: Adaptive temperature in {LLM} reinforcement learning.
\newblock In \emph{International Conference on Learning Representations (ICLR)}, 2026.
\newblock arXiv:2602.11779.

\bibitem[Fu et~al.(2025)Fu, Gao, Shen, Zhu, Mei, He, Xu, Wei, Mei, Wang, Yu, Yuan, and Wu]{fu2025areal}
Wei Fu, Jiaxuan Gao, Xujie Shen, Chen Zhu, Zhiyu Mei, Chuyi He, Shusheng Xu, Guo Wei, Jun Mei, Jiashu Wang, Tongkai Yu, Binhang Yuan, and Yi~Wu.
\newblock Areal: A large-scale asynchronous reinforcement learning system for language reasoning.
\newblock \emph{arXiv preprint arXiv:2505.24298}, 2025.

\bibitem[Hendrycks et~al.(2021)Hendrycks, Burns, Kadavath, Arora, Basart, Tang, Song, and Steinhardt]{hendrycks2021math}
Dan Hendrycks, Collin Burns, Saurav Kadavath, Akul Arora, Steven Basart, Eric Tang, Dawn Song, and Jacob Steinhardt.
\newblock Measuring mathematical problem solving with the {MATH} dataset.
\newblock In \emph{NeurIPS Datasets and Benchmarks Track}, 2021.

\bibitem[Huang et~al.(2026)Huang, Zhang, Hu, Yang, and Han]{huang2026vcpo}
Luke~J. Huang, Zhuoyang Zhang, Qinghao Hu, Shang Yang, and Song Han.
\newblock Stable asynchrony: Variance-controlled off-policy {RL} for {LLMs}.
\newblock \emph{arXiv preprint arXiv:2602.17616}, 2026.

\bibitem[Kwon et~al.(2023)Kwon, Li, Zhuang, Sheng, Zheng, Yu, Gonzalez, Zhang, and Stoica]{kwon2023vllm}
Woosuk Kwon, Zhuohan Li, Siyuan Zhuang, Ying Sheng, Lianmin Zheng, Cody~Hao Yu, Joseph~E. Gonzalez, Hao Zhang, and Ion Stoica.
\newblock Efficient memory management for large language model serving with {PagedAttention}.
\newblock In \emph{Proceedings of the 29th Symposium on Operating Systems Principles (SOSP)}, 2023.

\bibitem[Li et~al.(2027)Li, Lin, Fu, Zhou, Ji, Zhao, Wang, Jiang, and Cui]{stalenessconstrained2026}
Haoyang Li, Sheng Lin, Fangcheng Fu, Yuming Zhou, Xiaodong Ji, Yanfeng Zhao, Lefeng Wang, Jie Jiang, and Bin Cui.
\newblock {StaleFlow}: Staleness-aware data management for mitigating data skewness in fully disaggregated {RL} post-training.
\newblock In \emph{Proceedings of the ACM SIGMOD International Conference on Management of Data}, 2027.
\newblock arXiv:2601.12784.

\bibitem[Li et~al.(2025)Li, Wu, and Shen]{a3po2025}
Xiaocan Li, Shiliang Wu, and Zheng Shen.
\newblock {A-3PO}: Accelerating asynchronous {LLM} training with staleness-aware proximal policy approximation.
\newblock \emph{arXiv preprint arXiv:2512.06547}, 2025.

\bibitem[Lightman et~al.(2024)Lightman, Kosaraju, Burda, Edwards, Baker, Lee, Leike, Schulman, Sutskever, and Cobbe]{lightman2024verify}
Hunter Lightman, Vineet Kosaraju, Yuri Burda, Harrison Edwards, Bowen Baker, Teddy Lee, Jan Leike, John Schulman, Ilya Sutskever, and Karl Cobbe.
\newblock Let's verify step by step.
\newblock In \emph{International Conference on Learning Representations (ICLR)}, 2024.

\bibitem[Liu et~al.(2025)Liu, Diao, Hu, Lu, Dong, Zhang, Bukharin, Zhang, Zeng, Sreedhar, Shen, Mosallanezhad, Zhang, Yang, Yang, Kuchaiev, Liu, Yu, Molchanov, Choi, Kautz, and Dong]{liu2025prorl}
Mingjie Liu, Shizhe Diao, Jian Hu, Ximing Lu, Xin Dong, Hao Zhang, Alexander Bukharin, Shaokun Zhang, Jiaqi Zeng, Makesh~Narsimhan Sreedhar, Gerald Shen, David Mosallanezhad, Di~Zhang, Jonas Yang, June Yang, Oleksii Kuchaiev, Guilin Liu, Zhiding Yu, Pavlo Molchanov, Yejin Choi, Jan Kautz, and Yi~Dong.
\newblock Scaling up {RL}: Unlocking diverse reasoning in {LLMs} via prolonged training.
\newblock \emph{arXiv preprint arXiv:2507.12507}, 2025.

\bibitem[Lu et~al.(2025)Lu, Liu, Xiong, He, Gao, Wu, Wang, Liu, Li, Zhao, Huang, Yang, Li, Luo, Liu, Pan, Yan, Wang, Su, Wang, Qu, and Zheng]{rollflash2025}
Han Lu, Zichen Liu, Shaopan Xiong, Yancheng He, Wei Gao, Yanan Wu, Weixun Wang, Jiashun Liu, Yang Li, Haizhou Zhao, Ju~Huang, Siran Yang, Xiaoyang Li, Yijia Luo, Zihe Liu, Ling Pan, Junchi Yan, Wei Wang, Wenbo Su, Jiamang Wang, Lin Qu, and Bo~Zheng.
\newblock Part {II}: {ROLL} {Flash} -- accelerating {RLVR} and agentic training with asynchrony.
\newblock \emph{arXiv preprint arXiv:2510.11345}, 2025.

\bibitem[Shao et~al.(2024)Shao, Wang, Zhu, Xu, Song, Bi, Zhang, Zhang, Li, Wu, and Guo]{shao2024deepseekmath}
Zhihong Shao, Peiyi Wang, Qihao Zhu, Runxin Xu, Junxiao Song, Xiao Bi, Haowei Zhang, Mingchuan Zhang, Y.K. Li, Y.~Wu, and Daya Guo.
\newblock Deepseekmath: Pushing the limits of mathematical reasoning in open language models.
\newblock \emph{arXiv preprint arXiv:2402.03300}, 2024.

\bibitem[Sheng et~al.(2024)Sheng, Zhang, Ye, Wu, Zhang, Zhang, Peng, Lin, and Wu]{sheng2024verl}
Guangming Sheng, Chi Zhang, Zilingfeng Ye, Xibin Wu, Wang Zhang, Ru~Zhang, Yanghua Peng, Haibin Lin, and Chuan Wu.
\newblock Hybridflow: A flexible and efficient rlhf framework.
\newblock \emph{arXiv preprint arXiv:2409.19256}, 2024.

\bibitem[Song et~al.(2026)Song, Xu, Xiao, Bao, Shi, Feng, Wang, Han, Wu, Zhang, and Shi]{song2026staleness}
Jingwei Song, Haofeng Xu, Jie Xiao, Chengke Bao, Jingwei Shi, Pengbin Feng, Weixun Wang, Yuhang Han, Chuan Wu, Linfeng Zhang, and Bill Shi.
\newblock Staleness-learning rate scaling laws for asynchronous {RLHF}.
\newblock \emph{arXiv preprint arXiv:2607.01083}, 2026.

\bibitem[Tian et~al.(2026)Tian, Xie, and Wei]{mugrpo2026}
Minghao Tian, Yunfei Xie, and Chen Wei.
\newblock How off-policy can {GRPO} be? {Mu-GRPO} for efficient {LLM} reinforcement learning.
\newblock \emph{arXiv preprint arXiv:2605.17570}, 2026.

\bibitem[Xu et~al.(2026)Xu, Su, Tian, Diao, Qian, and Wu]{gac2026}
Haofeng Xu, Junwei Su, Yukun Tian, Lansong Diao, Zhengping Qian, and Chuan Wu.
\newblock {GAC}: Stabilizing asynchronous {RL} training for {LLMs} via gradient alignment control.
\newblock \emph{arXiv preprint arXiv:2603.01501}, 2026.

\bibitem[Yang et~al.(2024)Yang, Zhang, Hui, Gao, Yu, Li, Liu, Tu, Zhou, Lin, et~al.]{yang2024qwen25math}
An~Yang, Beichen Zhang, Binyuan Hui, Bofei Gao, Bowen Yu, Chengpeng Li, Dayiheng Liu, Jianhong Tu, Jingren Zhou, Junyang Lin, et~al.
\newblock {Qwen2.5-Math} technical report: Toward mathematical expert model via self-improvement.
\newblock \emph{arXiv preprint arXiv:2409.12122}, 2024.

\bibitem[Yang et~al.(2026)Yang, Yu, Wu, Sun, Wang, and Yang]{tsopsd2026}
Xuewei Yang, Jiachen Yu, Jie Wu, Shaoning Sun, Junjie Wang, and Yujiu Yang.
\newblock Internalize the temperature: On-policy self-distillation as policy reheater for reinforcement learning.
\newblock \emph{arXiv preprint arXiv:2606.00755}, 2026.

\bibitem[Yao et~al.(2025)Yao, Liu, Zhang, Dong, Shang, and Gao]{yao2025mismatch}
Feng Yao, Liyuan Liu, Dinghuai Zhang, Chengyu Dong, Jingbo Shang, and Jianfeng Gao.
\newblock On the rollout-training mismatch in modern {RL} systems.
\newblock In \emph{NeurIPS 2025 Workshop on Efficient Reasoning}, 2025.

\bibitem[Zheng et~al.(2026)Zheng, Zhao, and Chen]{zheng2026m2po}
Haizhong Zheng, Jiawei Zhao, and Beidi Chen.
\newblock Prosperity before collapse: How far can off-policy {RL} reach with stale data on {LLMs}?
\newblock In \emph{International Conference on Learning Representations (ICLR)}, 2026.
\newblock arXiv:2510.01161.

\bibitem[Zhou et~al.(2026)Zhou, Li, Cheng, Fan, and Cheng]{lookinward2026}
Yixiao Zhou, Yang Li, Dongzhou Cheng, Hehe Fan, and Yu~Cheng.
\newblock Look inward to explore outward: Learning temperature policy from {LLM} internal states via hierarchical {RL}.
\newblock \emph{arXiv preprint arXiv:2602.13035}, 2026.

\bibitem[Zhuang et~al.(2025)Zhuang, Zhou, Guo, Huang, Liu, Song, and Zhang]{multitemp2025}
Haomin Zhuang, Yujun Zhou, Taicheng Guo, Yue Huang, Fangxu Liu, Kai Song, and Xiangliang Zhang.
\newblock Exploring multi-temperature strategies for token- and rollout-level control in {RLVR}.
\newblock \emph{arXiv preprint arXiv:2510.08892}, 2025.

\end{thebibliography}
\bibliographystyle{iclr2027/iclr2027_conference}

\appendix
\section{Adaptive temperature control}
\label{app:adaptive}
\begin{figure}[h]
\centering
\includegraphics[width=\linewidth]{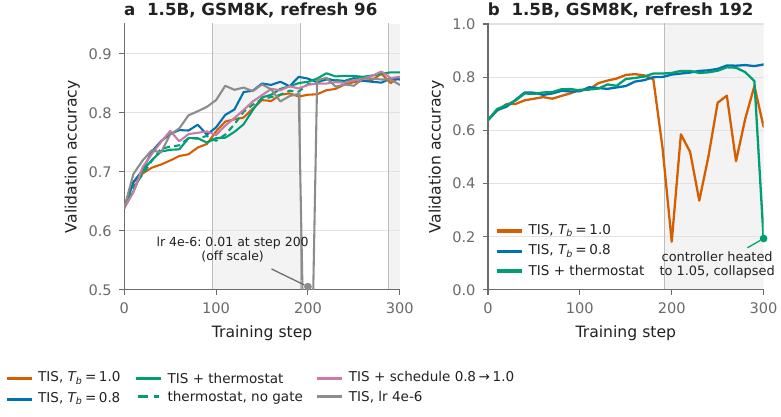}
\caption{Adaptive and scheduled temperature. (a) Refresh every 96: a saturation-triggered controller (with and without a KL gate) and a linear schedule from 0.8 to 1.0 all end within 0.01 of fixed cooling and of uncooled \tis; a doubled learning rate reaches 0.85 earlier but suffers one transient collapse at step 200. (b) Refresh every 192: the controller heats the sampler to 1.05 in the second cycle and collapses.}
\end{figure}
The controller keeps an exponential moving average of the fraction of groups with mixed rewards over 14 of the 16 prompts per step, and samples the remaining two prompts at $\Tb\pm0.1$ as probes. When the mixed fraction falls below 0.55 it moves $\Tb$ by 0.05 in the direction whose probe has more mixed groups, with a cooldown of eight steps, bounds $[0.6, 1.2]$, a reset to 0.8 at every refresh, and a gate that blocks moves away from 1.0 while the KL moving average exceeds 0.3. At $N{=}96$ the controller held 0.8 for the first cycle and heated to 1.0--1.05 in the later cycles; at $N{=}192$ it heated to 1.05 in the second cycle and the run collapsed at step 290--300 with a KL maximum of 0.43, below the gate threshold.

\section{All realizations of the 7B arms at refresh interval 144}
\label{app:7b}

\paragraph{The second cycle, aligned by cycle age.} Table~\ref{tab:cycle} reads each run at the same age within its first and second cycle (55--65 steps after the cycle start) and at the start of each cycle. In every cooled run except 7B seed 43, the learner--sampler mismatch at the same cycle age is lower in the second cycle than in the first, so later cycles are not generically harder; the seed-43 run is the one whose second cycle starts from the lowest entropy and whose mismatch then grows fastest. Whether the sharper refreshed policy drives the faster growth is a question for more runs; the uncooled runs are shown for contrast, where the refreshed sampler is already degenerate at the start of the second cycle.

\begin{table}[h]
\centering\footnotesize
\caption{Cycle-aligned readings. KL: learner--sampler KL averaged over cycle ages 55--65; entropy: learner entropy averaged over the first five steps of the cycle; clip: fraction of sampled responses at the length limit at ages 55--65. Cycle 1 / cycle 2.}
\label{tab:cycle}
\vspace{2pt}
\begin{tabular}{llccc}
\toprule
Run & $\Tb$ & KL & Entropy at start & Clip \\
\midrule
7B, $N{=}144$, seed 1 (full) & 0.8 & 0.134 / 0.068 & 0.27 / 0.16 & 0.19 / 0.01 \\
7B, $N{=}144$, seed 1 (cut) & 0.8 & 0.125 / 0.126 & 0.27 / 0.14 & 0.19 / 0.02 \\
7B, $N{=}144$, seed 43 & 0.8 & 0.080 / 0.218 & 0.28 / 0.11 & 0.19 / 0.03 \\
7B, $N{=}144$, seed 1 & 1.0 & 0.31 / 15.6 & 0.63 / 1.29 & 0.25 / 0.93 \\
7B, $N{=}144$, seed 43 & 1.0 & 0.32 / 23.1 & 0.63 / 0.03 & 0.28 / 0.95 \\
1.5B, $N{=}192$, 3 seeds & 0.8 & 0.03 / 0.01 & 0.28--0.30 / 0.17--0.19 & 0.14--0.16 / 0.05--0.08 \\
1.5B, $N{=}192$, 2 seeds & 1.0 & 0.05 / 4--8 & 0.74--0.96 / 0.31--0.35 & 0.25 / 0.39--0.81 \\
\bottomrule
\end{tabular}
\end{table}

\begin{figure}[H]
\centering
\includegraphics[width=0.8\linewidth]{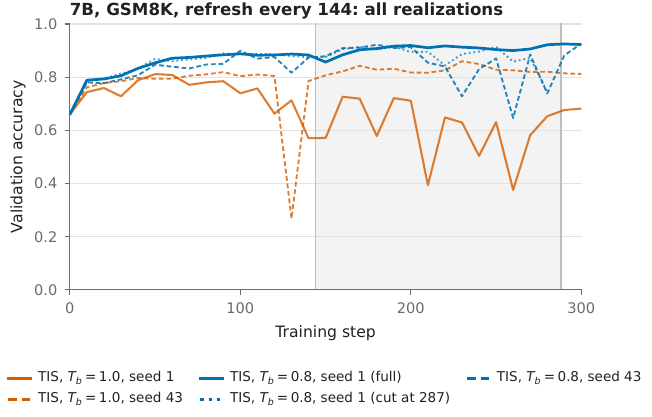}
\caption{Every run at refresh interval 144: two data seeds per arm, plus a first attempt at the cooled seed-1 arm that ended at step 287 (it matches the full run through the first refresh and drops at step 280). No cooled realization degrades before the first refresh; two of three show a late-second-cycle drop (seed 1 first attempt at 280; seed 43 from 230, KL above one nat and entropy rising, recovering at the refresh at 288). Both uncooled runs degrade severely before the first refresh, and after it their sampler produces length-clipped responses with training reward below 0.2 for the whole second cycle. All are listed in Table~\ref{tab:ledger}.}
\end{figure}

\section{Implementation}
\label{app:impl}
All experiments run on verl with the GRPO recipe of Section~\ref{sec:setup} and three small patches that do not change the loss, and a fourth that adds the $\mu$-GRPO loss of Section~\ref{sec:update}.

\paragraph{Staleness gate.} verl synchronises the learner's weights into the vLLM engine before every generation call. The gate skips that synchronisation except every $N$-th training call, and always synchronises before a validation call, so validation is on fresh weights while training samples come from a sampler that is up to $N-1$ steps old. Because the engine starts from the initial checkpoint, the first $N$ steps of every run sample from the initial policy.

\paragraph{Decoupled temperature.} verl applies one temperature to sampling and to the learner's log-probability computation. The patch reads a separate learner temperature (always $1$ in this paper) for the actor, the old-policy and the reference forward passes, and leaves the sampling temperature at $\Tb$. The behaviour log-probabilities that enter the importance weight are requested from vLLM in \texttt{processed\_logprobs} mode, so they are the probabilities of the tempered distribution that actually produced the tokens. With the default \texttt{raw\_logprobs} mode the recorded probabilities would be those of the untempered policy, and the weight $\pi_\theta/q$ would be wrong by a factor that grows with the sharpening; we verified on short runs that the two modes give different importance weights and that only the processed mode yields the behaviour of Section~\ref{sec:main}.

\paragraph{Per-prompt temperatures.} The controller of Appendix~\ref{app:adaptive} needs different prompts in one batch to be sampled at different temperatures. A third patch reads a per-prompt temperature list and clones the engine's sampling parameters per request; each data-parallel rank holds a contiguous block of prompts, so the mapping is by rank and position. We checked the plumbing by sampling half of a batch at temperature $0.05$ and half at $1.2$ and confirming near-duplicate responses in the first half only.

\paragraph{$\mu$-GRPO port.} For Section~\ref{sec:update} we registered a fourth policy loss in verl implementing the $\mu$-GRPO objective \citep{mugrpo2026} on top of the recipe above: $\rho=\exp(\log\pi_\theta-\log\beta)$ with $\beta$ the sampler's processed log-probabilities (a cooled sampler enters through its actual temperature-$\Tb$ distribution), surrogate $\min(\rho A,\,\mathrm{clip}(\rho,0,5)\,A)$, and a negative-advantage response containing a token with $\rho<10^{-4}$ removed from the loss through the mask. The truncated importance multiplier is off. Aggregation is the recipe's token mean and the KL penalty to the reference is kept, whereas the original normalises per response before averaging and trains a 7B model with a smaller learning rate and longer stages; the comparison is of the update rule under a common recipe, not a reproduction of the original system. The logged veto and clip fractions confirm that the rule is active.

\paragraph{Compute.} A 300-step 1.5B run takes about 3.2 hours on four 32~GB GPUs (MATH: 3.6 hours); a 7B run takes about 9 hours on four 96~GB GPUs. The 55 runs in Appendix~\ref{app:ledger} total roughly 220 GPU-box-hours.

\section{Full results ledger}
\label{app:ledger}
{\scriptsize\setlength{\tabcolsep}{3.2pt}
\begin{longtable}{lllllllrrrrrr}
\caption{Every 300-step run. Peak and final validation accuracy (GSM8K test or MATH-500), degradation onset ($\le$ running max $-0.10$), severe degradation ($\le$ max $-0.15$), first step with learner--sampler KL above 1 nat, and KL maximum. $T_b$: sampler temperature (* = initial value of a controller/schedule); $T_\ell$: learner temperature (tied = equal to $T_b$). Seed: data seed. `cut' = run interrupted; `re-run' = sampling replicate.}\label{tab:ledger}\\
\toprule Model & Task & $N$ & lr & Correction & $T_b$ & $T_\ell$ & Seed & Peak & Final & Onset & Severe & KL$>$1 \\ \midrule \endfirsthead
\toprule Model & Task & $N$ & lr & Correction & $T_b$ & $T_\ell$ & Seed & Peak & Final & Onset & Severe & KL$>$1 \\ \midrule \endhead
\bottomrule \endfoot
1.5B & GSM8K & 64 & 2e-6 & none & 1.0 & 1.0 & 1 & 0.822 & 0.723 & 120 & 190 & 159 \\
1.5B & GSM8K & 96 & 2e-6 & TIS & 0.8 & 1.0 & 1 & 0.861 & 0.856 & -- & -- & -- \\
1.5B & GSM8K & 96 & 2e-6 & TIS & 0.8 & tied & 1 & 0.864 & 0.864 & -- & -- & -- \\
1.5B & GSM8K & 96 & 2e-6 & TIS & 0.8 & 1.0 & 43 & 0.865 & 0.860 & -- & -- & -- \\
1.5B & GSM8K & 96 & 2e-6 & TIS & 0.8 & 1.0 & 44 & 0.867 & 0.865 & -- & -- & -- \\
1.5B & GSM8K & 96 & 2e-6 & TIS & 1.0 & 1.0 & 1 & 0.866 & 0.861 & -- & -- & -- \\
1.5B & GSM8K & 96 & 2e-6 & TIS & 1.0 & 1.0 & 43 & 0.860 & 0.860 & -- & -- & -- \\
1.5B & GSM8K & 96 & 2e-6 & TIS & 1.0 & 1.0 & 44 & 0.864 & 0.864 & -- & -- & -- \\
1.5B & GSM8K & 96 & 2e-6 & TIS + controller & 0.8* & 1.0 & 1 & 0.868 & 0.868 & -- & -- & -- \\
1.5B & GSM8K & 96 & 2e-6 & TIS + ctrl (no gate) & 0.8* & 1.0 & 1 & 0.862 & 0.862 & -- & -- & -- \\
1.5B & GSM8K & 96 & 2e-6 & TIS + schedule & 0.8* & 1.0 & 1 & 0.870 & 0.861 & -- & -- & -- \\
1.5B & GSM8K & 96 & 2e-6 & none & 0.8 & 1.0 & 1 & 0.853 & 0.029 & 70 & 190 & 184 \\
1.5B & GSM8K & 96 & 2e-6 & none & 0.8 & 1.0 & 43 & 0.813 & 0.354 & 140 & 140 & 93 \\
1.5B & GSM8K & 96 & 2e-6 & none & 0.8 & 1.0 & 44 & 0.826 & 0.290 & 130 & 140 & 133 \\
1.5B & GSM8K & 96 & 2e-6 & none, feedback cool & 0.8* & 1.0 & 1 & 0.867 & 0.867 & 160 & 160 & 158 \\
1.5B & GSM8K & 96 & 4e-6 & TIS & 1.0 & 1.0 & 1 & 0.859 & 0.847 & 200 & 200 & 96 \\
1.5B & GSM8K & 192 & 1.5e-6 & TIS & 1.0 & 1.0 & 43 & 0.823 & 0.820 & -- & -- & -- \\
1.5B & GSM8K & 192 & 1.5e-6 & TIS & 1.0 & 1.0 & 44 & 0.837 & 0.829 & -- & -- & -- \\
1.5B & GSM8K & 192 & 1e-6 & TIS & 1.0 & 1.0 & 43 & 0.798 & 0.798 & -- & -- & -- \\
1.5B & GSM8K & 192 & 1e-6 & TIS & 1.0 & 1.0 & 44 & 0.794 & 0.792 & -- & -- & -- \\
1.5B & GSM8K & 192 & 2e-6 & $\mu$-GRPO & 0.8 & 1.0 & 43 & 0.831 & 0.831 & -- & -- & -- \\
1.5B & GSM8K & 192 & 2e-6 & $\mu$-GRPO & 0.8 & 1.0 & 44 & 0.827 & 0.825 & -- & -- & -- \\
1.5B & GSM8K & 192 & 2e-6 & $\mu$-GRPO & 1.0 & 1.0 & 43 & 0.839 & 0.835 & -- & -- & -- \\
1.5B & GSM8K & 192 & 2e-6 & $\mu$-GRPO & 1.0 & 1.0 & 44 & 0.852 & 0.852 & -- & -- & -- \\
1.5B & GSM8K & 192 & 2e-6 & TIS & 0.8 & 1.0 & 1 & 0.848 & 0.848 & -- & -- & -- \\
1.5B & GSM8K & 192 & 2e-6 & TIS & 0.8 & tied & 1 & 0.843 & 0.843 & -- & -- & -- \\
1.5B & GSM8K & 192 & 2e-6 & TIS & 0.8 & 1.0 & 43 & 0.866 & 0.857 & -- & -- & -- \\
1.5B & GSM8K & 192 & 2e-6 & TIS & 0.8 & 1.0 & 44 & 0.859 & 0.859 & -- & -- & -- \\
1.5B & GSM8K & 192 & 2e-6 & TIS & 0.9 & 1.0 & 1 & 0.855 & 0.842 & -- & -- & -- \\
1.5B & GSM8K & 192 & 2e-6 & TIS & 1.0 & 1.0 & 1 & 0.811 & 0.613 & 190 & 190 & 179 \\
1.5B & GSM8K & 192 & 2e-6 & TIS & 1.0 & 1.0 & 43 & 0.796 & 0.588 & 190 & 190 & 171 \\
1.5B & GSM8K & 192 & 2e-6 & TIS & 1.0 & 1.0 & 44 & 0.826 & 0.469 & 190 & 190 & 181 \\
1.5B & GSM8K & 192 & 2e-6 & TIS + controller & 0.8* & 1.0 & 1 & 0.837 & 0.193 & 300 & 300 & -- \\
1.5B & GSM8K & 288 & 2e-6 & TIS & 0.6 & 1.0 & 1 & 0.837 & 0.564 & 300 & 300 & 221 \\
1.5B & GSM8K & 288 & 2e-6 & TIS & 0.7 & 1.0 & 1 & 0.813 & 0.745 & 200 & 200 & 216 \\
1.5B & GSM8K & 288 & 2e-6 & TIS & 0.7 & 1.0 & 1 (cut) & 0.825 & 0.603 & 190 & 200 & 217 \\
1.5B & GSM8K & 288 & 2e-6 & TIS & 0.8 & 1.0 & 1 & 0.845 & 0.845 & 250 & 250 & 245 \\
1.5B & GSM8K & 288 & 2e-6 & TIS & 1.0 & 1.0 & 1 & 0.807 & 0.786 & 190 & 190 & 179 \\
1.5B & MATH & 96 & 2e-6 & TIS & 1.0 & 1.0 & 1 & 0.688 & 0.678 & -- & -- & -- \\
1.5B & MATH & 96 & 2e-6 & none & 0.8 & 1.0 & 1 & 0.706 & 0.706 & 200 & -- & 176 \\
1.5B & MATH & 192 & 2e-6 & TIS & 0.8 & 1.0 & 1 & 0.682 & 0.682 & -- & -- & -- \\
1.5B & MATH & 192 & 2e-6 & TIS & 0.8 & 1.0 & 1 (cut) & 0.680 & 0.680 & -- & -- & -- \\
1.5B & MATH & 192 & 2e-6 & TIS & 1.0 & 1.0 & 1 & 0.698 & 0.698 & -- & -- & -- \\
1.5B & MATH & 288 & 4e-6 & TIS & 0.6 & 1.0 & 1 & 0.714 & 0.223 & 160 & 160 & 164 \\
1.5B & MATH & 288 & 4e-6 & TIS & 0.8 & 1.0 & 1 & 0.698 & 0.414 & 260 & 260 & 247 \\
1.5B & MATH & 288 & 4e-6 & TIS & 0.8 & 1.0 & 43 & 0.700 & 0.386 & 200 & 280 & 215 \\
1.5B & MATH & 288 & 4e-6 & TIS & 1.0 & 1.0 & 1 & 0.680 & 0.000 & 180 & 180 & 171 \\
1.5B & MATH & 288 & 4e-6 & TIS & 1.0 & 1.0 & 43 & 0.692 & 0.590 & 200 & 200 & 172 \\
7B & GSM8K & 144 & 2e-6 & TIS & 0.8 & 1.0 & 1 & 0.914 & 0.745 & 280 & 280 & -- \\
7B & GSM8K & 144 & 2e-6 & TIS & 0.8 & 1.0 & 1 (re-run) & 0.925 & 0.923 & -- & -- & -- \\
7B & GSM8K & 144 & 2e-6 & TIS & 0.8 & 1.0 & 43 & 0.926 & 0.926 & 230 & 230 & 232 \\
7B & GSM8K & 144 & 2e-6 & TIS & 1.0 & 1.0 & 1 & 0.811 & 0.682 & 120 & 140 & 98 \\
7B & GSM8K & 144 & 2e-6 & TIS & 1.0 & 1.0 & 43 & 0.860 & 0.812 & 130 & 130 & 88 \\
7B & GSM8K & 192 & 2e-6 & TIS & 0.8 & 1.0 & 1 & 0.911 & 0.873 & 180 & 180 & 148 \\
7B & GSM8K & 192 & 2e-6 & TIS & 1.0 & 1.0 & 1 & 0.860 & 0.000 & 120 & 120 & 92 \\
\end{longtable}
}

\end{document}